\documentclass[conference]{IEEEtran}
\IEEEoverridecommandlockouts

\usepackage{cite}
\ifCLASSINFOpdf
\else
\fi
\usepackage{microtype}
\usepackage{array}
\usepackage{mdwmath}
\usepackage{mdwtab}
\usepackage{graphicx}
\usepackage{euscript}
\usepackage{amsfonts}
\usepackage{bigstrut}
\usepackage{tabularx}
\usepackage{cite}
\usepackage{amsmath}
\usepackage{algorithm}
\usepackage[noend]{algpseudocode}
\makeatletter
\def\BState{\State\hskip-\ALG@thistlm}
\makeatother
\usepackage{enumitem}
\usepackage{upgreek}
\usepackage{dsfont}
\usepackage{amsthm}
\usepackage{amssymb}
\usepackage{textcomp}
\usepackage{gensymb}
\usepackage{xcolor}
\usepackage{mathtools}

\usepackage{hyperref}
\usepackage{epstopdf} 
\usepackage{bbm}
\usepackage{microtype}
\usepackage{booktabs}

\usepackage{multirow}
\usepackage{threeparttable}  
\def\BibTeX{{\rm B\kern-.05em{\sc i\kern-.025em b}\kern-.08em
    T\kern-.1667em\lower.7ex\hbox{E}\kern-.125emX}}
\begin{document}

\title{Resource-Efficient Multiview Perception: Integrating Semantic Masking with Masked Autoencoders}



\author{Kosta {Dakic}$^\dagger$, Kanchana Thilakarathna$^\dagger$, Rodrigo N. Calheiros$^\star$ and Teng Joon Lim$^\dagger$\\
	$^\dagger$ Faculty of Engineering, The University of Sydney, Australia\\
    $^\star$ School of Computer, Data \& Mathematical Sciences, Western Sydney University, Australia\\
Emails: kosta.dakic@ieee.org, tj.lim@sydney.edu.au
}

\maketitle

\begin{abstract}
Multiview systems have become a key technology in modern computer vision, offering advanced capabilities in scene understanding and analysis. However, these systems face critical challenges in bandwidth limitations and computational constraints, particularly for resource-limited camera nodes like drones. This paper presents a novel approach for communication-efficient distributed multiview detection and tracking using masked autoencoders (MAEs). We introduce a semantic-guided masking strategy that leverages pre-trained segmentation models and a tunable power function to prioritize informative image regions. This approach, combined with an MAE, reduces communication overhead while preserving essential visual information. We evaluate our method on both virtual and real-world multiview datasets, demonstrating comparable performance in terms of detection and tracking performance metrics compared to state-of-the-art techniques, even at high masking ratios. Our selective masking algorithm outperforms random masking, maintaining higher accuracy and precision as the masking ratio increases. Furthermore, our approach achieves a significant reduction in transmission data volume compared to baseline methods, thereby balancing multiview tracking performance with communication efficiency.
\end{abstract}
\begin{IEEEkeywords}
Multiview, Semantic Masking, Masked Autoencoders, Communication-Efficient, Distributed Vision
\end{IEEEkeywords}

\section{Introduction}~\label{Section_I}
Multiview systems play an increasingly important role in modern computer vision and pervasive computing, providing enhanced capabilities for comprehensive scene interpretation. These systems leverage the collaboration of multiple cameras to capture diverse perspectives of a scene, enabling a more thorough and robust interpretation of complex environments~\cite{1233912}. The applications of such systems span a wide spectrum, including but not limited to urban surveillance, disaster response coordination, intelligent traffic monitoring, and advanced sports analytics~\cite{9200313}. By integrating data from various viewpoints, multiview systems can be effective in addressing challenges such as occlusions, restricted fields of view, varying illumination conditions, objects with flexible or changing shapes, and scenarios involving extended periods of concealment~\cite{otsurvey}.

In the context of pervasive computing, these multiview systems represent a significant step towards creating smart environments where computational power and sensing capabilities are seamlessly integrated into the physical world~\cite{7807196}. By distributing cameras throughout an environment, we create ubiquitous sensing networks that provide rich, contextual information about spaces and their occupants. However, the widespread deployment of these pervasive multiview systems exacerbates existing challenges and introduces new ones. As the number of cameras increases, the volume of visual data that needs to be transmitted and processed becomes excessive, straining network resources, especially in bandwidth-limited scenarios. Moreover, the real-time processing of high-dimensional video streams demands substantial computational power, which is particularly challenging for resource-limited camera nodes like drones or satellites~\cite{10286330}.

In this context, the Masked Autoencoder~\cite{9879206} (MAE) is proving to be an excellent tool for the reconstruction of masked regions of an image~\cite{10316537}. MAEs leverage the power of transformers and the reconstruction capabilities of autoencoders to process images and reconstruct data from partial observations. This makes them well-suited for enhancing resource efficiency, adaptability, and robustness in various wireless communication tasks~\cite{8054694}. Masking and transmitting only a subset of the data patches at the camera node reduces the communication overhead, but the edge server can still reconstruct the complete visual data using the MAE network.

Multiview target detection and tracking, a cornerstone application of multiview systems, involves the complex task of identifying and following objects or individuals across multiple camera views. This field has seen significant advancements in recent years, driven by the increasing demand for robust surveillance and monitoring solutions~\cite{9289536}. The challenge lies not only in accurately detecting targets from diverse angles but also in maintaining consistent object identities across different views, a problem known as cross-camera re-identification (Re-ID). Cutting-edge methods typically utilize a 2D to 3D projection, the camera views are aggregated, and the multiple targets are identified~\cite{9879609, Cheng_2023_ICCV}. Projecting camera views has revolutionized this domain, offering improved precision and accuracy in complex scene dynamics~\cite{BEVformer, hou2020multiview}. For instance, methods leveraging convolutional neural networks (CNNs) have shown remarkable success in extracting discriminative features for both detection and re-identification tasks~\cite{teepe2023earlybird}. Furthermore, the integration of attention mechanisms and transformer architectures has enabled more effective modeling of spatial-temporal relationships in multiview scenarios~\cite{BEVformer}. 

Despite these advancements, the trade-off between computational complexity and accuracy remains a critical consideration, especially for deployments with limited resources such as low-power camera nodes~\cite{9557791} and drone-based systems~\cite{9298794}. As the field progresses, there is a growing emphasis on developing more efficient algorithms that can leverage complementary information from multiple views while minimizing computational overhead and communication requirements. This intersection of multiview geometry, deep learning, and distributed computing presents both challenges and opportunities in the practical adoption of advanced multiview surveillance and monitoring applications.

In this paper, we propose a novel approach for efficient multiview detection and tracking using a semantic-guided masking strategy and MAE-based reconstruction. Our method comprises (i) Partially masking visual data from distributed cameras using a semantic-guided masking strategy to select and transmit only the most informative image patches. (ii) An edge server that reconstructs complete visual data for each camera view using the MAE encoder-decoder network. (iii) Multiview fusion of the reconstructed data to generate a comprehensive scene representation. (iv) CNN-based processing of the fused data for detection and tracking. This approach achieves efficient communication between cameras and the edge server while maintaining detection and tracking performance.

The main contributions of this work are as follows:
\begin{itemize}
\item Development of a novel semantic-guided masking technique and MAE-based reconstruction for multiview systems. This technique leverages pre-trained segmentation models and a tunable power function to prioritize informative image regions, enhancing detection and tracking performance while reducing communication overhead.
\item Design of a distributed multiview system architecture that balances computational load between camera nodes and the central edge server. This system demonstrates the trade-offs between communication volume, computational complexity at camera nodes, and overall detection and tracking performance.
\item Comprehensive evaluation and quantification of the performance-communication trade-off in multiview detection and tracking scenarios. Our analysis compares random masking versus semantic-guided masking across various masking ratios, providing insights into the efficacy of our approach in bandwidth-constrained environments.
\end{itemize}

The rest of this paper is organized as follows; Sec.~\ref{Section_II} covers the background literature, Sec.~\ref{Section_III} covers the system model, Sec.~\ref{Section_IV} shows the results, and Sec.~\ref{Section_V} concludes the research work.

\section{Background and Related work}~\label{Section_II}
\subsection{Target Detection and Tracking}
Recent advancements in object detection and tracking have seen a shift from traditional methods to deep learning approaches. Notable developments include the enhancement of the SORT algorithm~\cite{8296962}, the introduction of transformer-based models like DETR~\cite{carion2020end} and its deformable variant~\cite{zhu2021deformable}, and end-to-end trainable models for simultaneous detection and tracking~\cite{9879668}. For a comprehensive review, see Luo et al.~\cite{LUO2021103448}.

Multi-camera tracking introduces additional challenges, particularly in maintaining object identities across views. Ristani and Tomasi~\cite{8578730} proposed using CNNs for robust feature learning in Multi-Target Multi-Camera Tracking and Person Re-Identification. Addressing communication efficiency, Hu et al.~\cite{hu2022wherecomm} proposed Where2comm, a collaborative perception framework that uses spatial confidence maps to guide pragmatic compression and effective information aggregation. However, while Where2comm focuses on collaborative perception and bandwidth adaptation, it does not specifically address the computational constraints of camera nodes or the challenges of multiview detection and tracking. For a comprehensive review of recent advancements in Multi-target Multi-camera detection and tracking, we refer readers to the survey paper by Amosa et al.~\cite{AMOSA2023126558}.

\subsubsection{Multiview}
While multi-camera systems simply employ multiple cameras to capture a scene from different angles, multiview systems go a step further by integrating and processing data from these multiple perspectives to create a unified understanding of the environment. Multiview detection and tracking have seen significant advancements in recent years, particularly in addressing challenges related to occlusion, crowdedness, and computational efficiency. Fleuret et al.~\cite{4359319} laid the groundwork with their probabilistic occupancy map (POM) approach, demonstrating robust multi-person detection in complex environments. Subsequent works have focused on leveraging bird's eye view (BEV) representations to improve detection and tracking performance. Hou et al.~\cite{hou2020multiview} introduced MVDet, which projects features from multiple camera views onto a shared BEV representation, demonstrating the effectiveness of anchor-free representation and fully convolutional spatial aggregation. Their follow-up work, MVDeTr~\cite{101145}, incorporated a shadow transformer to handle projection distortions more effectively. Building on these concepts, Teepe et al.~\cite{teepe2023earlybird} proposed EarlyBird, which performs early fusion of camera views in BEV space and introduces re-identification features for appearance-based tracking. Their subsequent work~\cite{teepe2024lifting} explored various lifting algorithms for projecting multiview features into BEV space, highlighting the importance of temporal aggregation.

While these approaches have improved multiview detection and tracking accuracy, they generally overlook the critical aspects of communication and computational efficiency, particularly in distributed systems with resource-constrained camera nodes. The CenterCoop framework by Zhou et al.~\cite{zhou2024centercoop} represents a step towards addressing communication efficiency in collaborative perception, but it does not fully consider the computational constraints of camera nodes. The OCMCTrack framework~\cite{Specker_2024_CVPR} comes closest to addressing the communication efficiency problem by separating single-camera processing from cross-camera association. However, it still requires significant computational resources at the camera for object detection, feature extraction, and single-camera tracking. This approach, while reducing communication costs, could be detrimental to battery-operated devices and requires relatively powerful computational platforms at the camera node.

Our proposed method uniquely addresses both communication and computational efficiency challenges that are largely overlooked in the existing literature. By employing a masking technique at the camera node, we reduce both the communication overhead and the computational burden on camera nodes. Unlike OCMCTrack, which performs complex computations at the camera node, our approach moves the bulk of the processing to the edge server, allowing for the use of simpler, more energy-efficient camera nodes. This is particularly crucial for battery-operated systems like drones or satellite networks. 

\subsection{Semantic Communications}

Semantic communication networks aim to convey the meaning of messages rather than focusing solely on accurate data transmission. This concept traces back to the seminal 1948 work~\cite{6773024}, which identified technical, semantic, and effectiveness levels in communication problems. Recent advancements in machine learning and wireless communications have made semantic communication increasingly feasible. A comprehensive framework for next-generation semantic communication networks~\cite{10554663} emphasizes minimalist, generalizable, and efficient semantic representations. This aligns with our proposed method, which uses a masking technique to compress visual data from distributed cameras, transmitting only essential information for efficient target detection and tracking.

Research in semantic communication spans various applications. An end-to-end deep learning system for text transmission, DeepSC~\cite{9398576}, was later extended to speech signals with DeepSC-S~\cite{9500590}. Further advancements include multi-modal and multi-user scenarios~\cite{9830752} and efforts to reduce model complexity~\cite{9252948}. In image transmission, relevant to our work, approaches include joint source-channel coding~\cite{8723589,9066966}, deep reinforcement learning for aerial image transmission~\cite{9796572}, and MAE networks to combat semantic noise~\cite{10101778}. Federated learning has also been applied for efficient distributed image transmission in IoT devices~\cite{10216933}.

Our work builds upon the foundations of MAE-based image transmission~\cite{10333576} but diverges from traditional semantic communication by introducing a semantically guided approach. We integrate semantic segmentation to identify regions of interest (ROI), enabling more efficient allocation of communication resources. While this aligns with recent AI-driven communication strategies for bandwidth-constrained scenarios~\cite{10100737}, our method is specifically tailored for multiview detection and tracking applications. 


\section{System Model}~\label{Section_III}
In this section, we present the proposed framework for our communication-efficient multiview system. We introduce the key components of the system, including the MAE and the perspective transformation for aggregating the distributed image sequences at the edge server.

\subsection{Framework}

\begin{figure*}[!t]
    \normalsize
	\centering
   	\includegraphics[width=\linewidth]{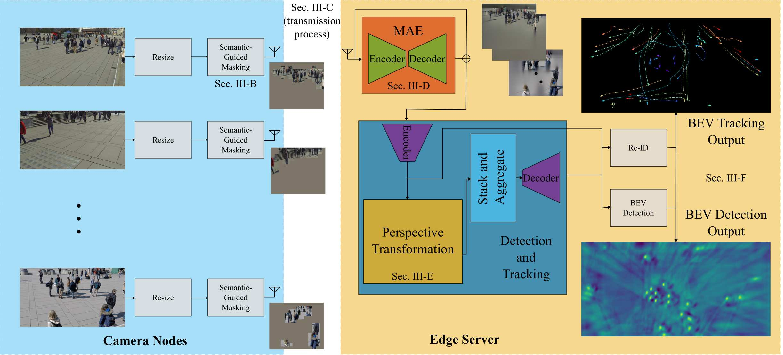}
	\caption{System model of the proposed distributed multiview pedestrian detection and tracking framework. Multiple cameras capture images, which are resized and undergo semantic-guided masking before wireless transmission. The received masked images are processed by an MAE with an encoder-decoder architecture. The reconstructed images are then fed into a combined detection and tracking module~\cite{hou2020multiview, teepe2023earlybird}, producing both detection and tracking outputs for the observed targets.}
	\label{Fig_system_model}
\end{figure*}

The proposed communication-efficient multiview system consists of multiple cameras (the camera nodes) that capture image sequences from different viewpoints positioned to have an overlapping field of view. The system aims to efficiently transmit the information present in the distributed image sequences to the edge server for target detection and tracking tasks.

The framework of the proposed system is illustrated in Fig.~\ref{Fig_system_model}. As an initial step, the images captured by the distributed cameras are resized to a lower resolution. This resizing serves two crucial purposes: firstly, it reduces the communication cost by decreasing the amount of data that needs to be transmitted; secondly, it addresses the computational complexity issue, as the processing time of MAEs increases quadratically with image resolution~\cite{9879206}. After resizing, the distributed cameras apply a masking process in preparation for the MAE e.g., they can perform computationally efficient random masking. Alternatively, we can employ semantically-guided masking (further explained in Sec.~\ref{sem}) at the camera nodes to improve performance at the cost of higher computational complexity as a semantic segmentation network needs to be used. At the edge server, the MAE first encodes the received masked images and then decodes the encoded latent data to output full images, filling in the masked patches (see Sec.~\ref{mae}). These reconstructed images are then aggregated using perspective transformation (explained in Sec.~\ref{perspective}) to obtain a unified view of the scene in the BEV. The aggregated view is subsequently used for target detection and tracking tasks (see Sec.~\ref{detntrack}).

\subsection{Semantic-Guided Masking} \label{sem}

While the idea of utilizing semantic information in masking~\cite{NEURIPS2022_5c186016} has shown promise, our approach offers several key innovations. Unlike previous methods that rely on custom part learning or gradual masking strategies, we leverage a state-of-the-art pre-trained semantic segmentation model to obtain robust and accurate semantic information across diverse image types. Our technique introduces a novel heatmap-based patch selection process, controlled by a tunable power function, which allows for fine-grained balancing of local and global information during the masking process.

In our proposed communication-efficient multiview system, we employ a selective masking strategy to prioritize the transmission of the most informative patches of the image to the edge server. This strategy is based on the premise that not all patches in an image are equally important for the target detection and tracking tasks. By selectively masking the less informative patches and transmitting only the most relevant ones, we can reduce the amount of data transmitted while preserving the essential semantic information.

The selective masking process begins by passing the input image through a pre-trained semantic segmentation network to obtain the semantic masks of the targets. In the context of our research, we focus on pedestrian detection in a multiview dataset. Therefore, we utilize a pre-trained Detectron2 network~\cite{wu2019detectron2} for semantic segmentation. Detectron2 is a well-established and highly accurate framework for object detection and segmentation tasks. By leveraging a pre-trained model, we can avoid the need to train our semantic segmentation network from scratch, as semantic segmentation is a well-researched problem and can be considered largely solved for our purposes.

Once the semantic masks of the pedestrians are obtained, we proceed to create a heatmap of the image that highlights the most ``active" patches. To do this, we first divide the image into patches of a fixed size (e.g., 20x20 pixels). For each patch, we calculate its activity level by counting the number of pixels belonging to the pedestrian masks within the patch itself and its eight neighboring patches. The sum of these pixel counts serves as a measure of the patch's activity level.

After calculating the activity levels for all patches, we apply a power function with a hyperparameter $\kappa$ to the activity levels. The power function is defined as $f(x) = x^\kappa$, where $x$ represents the activity level of a patch. The purpose of the power function is to control the randomness of the patch selection process. By adjusting the value of $\kappa$, we can influence the probability distribution of the patches based on their activity levels. A lower value of $\kappa$ increases the randomness, allowing patches with lower activity levels to have a higher chance of being selected, while a value of $\kappa$ closer to 1 emphasizes the importance of patches with higher activity levels. Once the power function is applied, we normalize the modified activity levels to obtain a probability distribution over the patches. We then randomly select a subset of patches to be kept unmasked based on this probability distribution. The number of patches to be kept unmasked is determined by a predefined masking ratio, which represents the percentage of patches to be transmitted.

\begin{figure*}[!t]
    \normalsize
	\centering
   	\includegraphics[width=\linewidth]{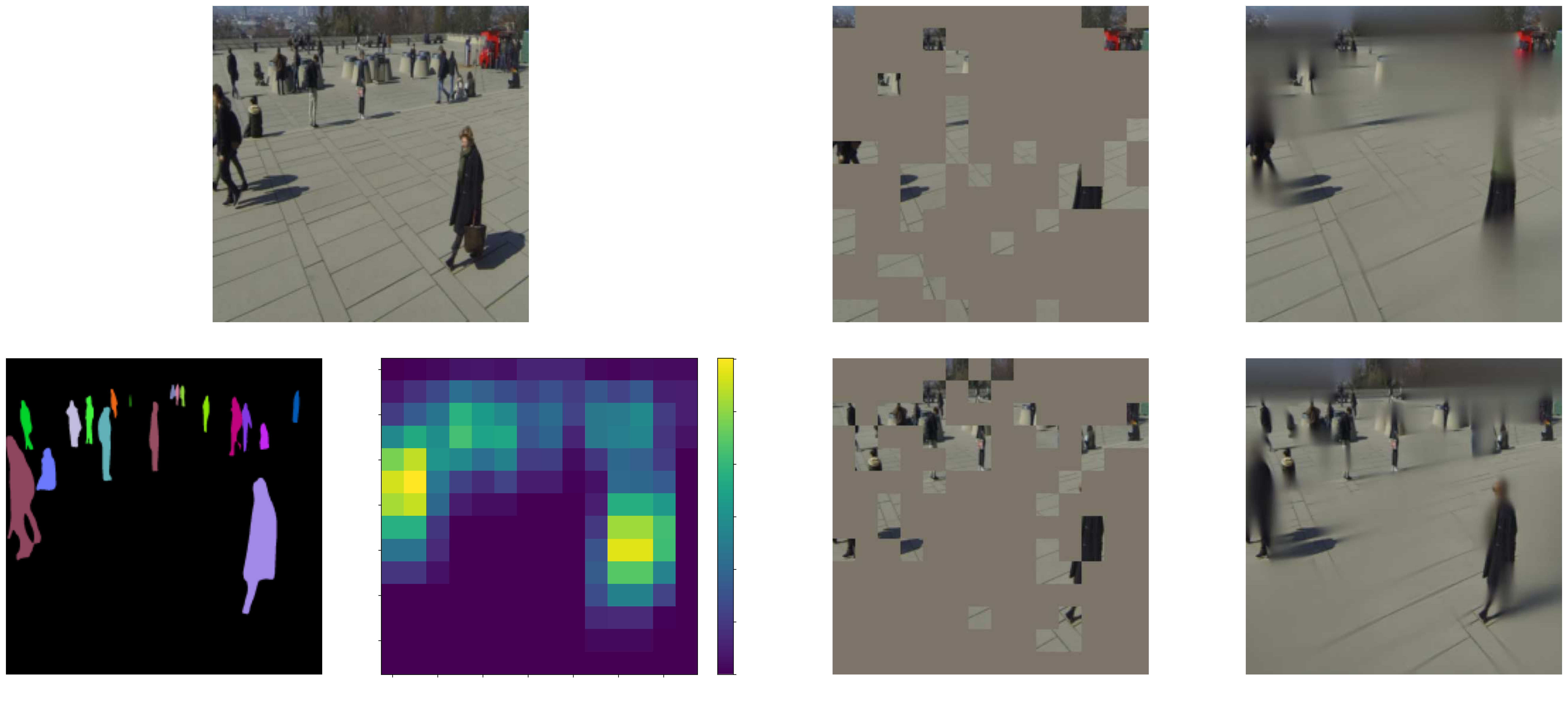}
	\caption{Visual comparison of random and semantically-guided masking techniques for image reconstruction using an MAE. The masking ratio is set at 0.8, and the $\kappa = 0.1$ for semantically-guided masking. (a) Original input image. (b) Randomly masked image. (c) Reconstruction from random masking. (d) Instance segmentation of the original image. (e) Heatmap of patch importance based on instance segmentation results. (f) semantically-guided masked image. (g) Reconstruction from semantically-guided masking. The semantically-guided masking technique (d-g) demonstrates improved reconstruction quality compared to random masking (b-c), particularly in areas of high semantic importance as indicated by the heatmap (e).}
	\label{Fig_mae_comp}
\end{figure*}

By selectively masking the less informative patches, we can reduce the amount of data that needs to be transmitted to the edge server. This approach allows us to focus on transmitting the most relevant information for the target detection and tracking tasks while minimizing the communication overhead. The selective masking strategy helps to strike a balance between data compression and the preservation of essential information. Fig.~\ref{Fig_mae_comp} shows a visual comparison of random masking compared to our semantically-guided masking technique.

\begin{algorithm}
\caption{Semantic-Guided Masking}
\label{Alg_preferential}
\begin{algorithmic}[1]
\Require Input image $I$, patch size $p$, masking ratio $r$, power parameter $\kappa$
\Ensure Masked image $I_{masked}$
\State $M \gets \text{SemanticSegmentation}(I)$ \Comment{Get semantic masks}
\State $P \gets \text{DivideIntoPatches}(I, p)$ \Comment{Divide image into patches}
\For{each patch $i$ in $P$}
    \State $A[i] \gets \sum_{j \in N(i)} \text{CountMaskPixels}(M, j)$ \Comment{Activity level}
\EndFor
\State $A' \gets A^\kappa$ \Comment{Apply power function}
\State $P \gets \text{Normalize}(A')$ \Comment{Normalize to get probability distribution}
\State $n \gets \lfloor |P| \cdot (1-r) \rfloor$ \Comment{Number of patches to keep}
\State $U \gets \text{RandomSample}(P, n, P)$ \Comment{Select unmasked patches}
\State $I_{masked} \gets \text{ApplyMask}(I, U)$
\Return $I_{masked}$
\end{algorithmic}
\end{algorithm}

Algorithm~\ref{Alg_preferential} performs semantic-guided masking by calculating patch activity levels based on semantic segmentation, applying a power function, normalizing to obtain a probability distribution, and randomly sampling patches to keep unmasked. The activity level calculation for each patch $i$ in step 4 is formulated as 
\begin{equation}
A[i] = \sum_{j \in N(i)} \sum_{x,y \in j} M(x,y),
\end{equation}
where $N(i)$ is the set of neighboring patches including $i$ itself, and $M(x,y)$ is the binary mask value at pixel $(x,y)$. Normalization to obtain probability distribution from step 6 is shown as 
\begin{equation}
P[i] = \frac{A'[i]}{\sum_j A'[j]}.
\end{equation}
Random sampling of unmasked patches from step 8 is formulated as 
\begin{equation}
U \sim \text{Multinomial}(n, P),
\end{equation}
where $n = \lfloor |P| \cdot (1-r) \rfloor$ is the number of patches to keep unmasked.

\subsection{Image Transmission Process} \label{transmission}

Our system begins by sending a seed for the masking pattern from each camera to the edge server. The number of patches determines this seed within an image, which in turn depends on the original image's size and the patches' size.

The number of patches in the image, $N$, is given by
\begin{equation}
    N = \left\lfloor\frac{W}{p}\right\rfloor \times \left\lfloor\frac{H}{p}\right\rfloor,
\end{equation}
where $W$ and $H$ are the width and height of the original image in pixels, $p$ is the size of each square patch in pixels, and $\lfloor \cdot \rfloor$ denotes the floor function.

The number of seeds required, $S$, is equal to the number of unmasked patches
\begin{equation}
    S = \lceil N \times (1 - r) \rceil ,
\end{equation}
where $r$ is the masking ratio (percentage of patches to be masked) and $\lceil \cdot \rceil$ denotes the ceiling function.
The idea after the seed is sent is to sequentially send over only the RGB values of the unmasked patches, overall, reducing the communication cost.

\subsection{MAE} \label{mae}
The MAE is a key component of the proposed system. It is based on the concept of masked image modeling, where a portion of the input image is masked, and the autoencoder is trained to reconstruct the original image from the masked input.

The architecture of the MAE consists of an encoder and a decoder. The encoder takes the masked image as input and extracts a compact feature representation. The decoder then reconstructs the original image from the feature representation. By training the MAE to reconstruct the original image from a masked input, it learns to capture the essential semantic information present in the image.

The masking strategy employed is designed to focus on the regions of the image that are most relevant for the target detection and tracking tasks. This is achieved by masking out the background regions of the image and preserving the foreground regions that are likely to contain the targets of interest. This masking strategy helps to reduce the amount of irrelevant information transmitted to the edge server, thereby improving the efficiency of the system.

During the training phase, the MAE is trained on a large dataset of images to learn a generic feature representation. For training, a random masking strategy is employed to ensure the model learns to reconstruct diverse image patterns. Once trained, the full MAE (both encoder and decoder) is deployed on the edge server. The camera nodes (distributed cameras) are responsible only for capturing images and applying the masking strategy.

In operation, each camera captures an image sequence and applies masking and only these unmasked patches are then transmitted to the edge server, reducing the data volume sent over the network. At the edge server, the received patches are first processed by the MAE encoder to extract features, and then the MAE decoder reconstructs the full images by filling in the masked regions based on these features.

This architecture minimizes the computational burden on the camera nodes, as they only perform image capture and masking. The more computationally intensive tasks of encoding and reconstruction are centralized at the server, which typically has more processing power. This approach allows for efficient use of network bandwidth while maintaining the ability to reconstruct high-quality images for subsequent detection and tracking tasks.

\subsection{Perspective Transformation} \label{perspective}
The perspective transformation occurs in the MVdet framework. It is used to aggregate the feature representations after the CNN decoder. The goal of the perspective transformation is to project the features from the different camera views onto a common reference plane, typically the BEV. The perspective transformation uses a homography matrix, which describes the mapping between the coordinate systems of the different camera views and the common reference plane. The homography matrix is estimated using corresponding points between the camera views, which can be obtained through camera calibration or feature-matching techniques. Once the homography matrix is estimated, the feature representations from the different camera views are projected onto the common reference plane using the perspective transformation. This results in a unified representation of the scene in the BEV, which facilitates the target detection and tracking tasks~\cite{hou2020multiview, teepe2023earlybird}.

The aggregated feature representation in the BEV provides a comprehensive view of the scene, allowing for more accurate and robust target detection and tracking compared to using the individual camera views separately. The perspective transformation helps to mitigate the effects of occlusions and viewpoint variations, as the targets of interest are likely to be visible in at least one of the camera views. The perspective transformation can be mathematically described using the pinhole camera model. The transformation between 3D world coordinates $(x, y, z)$ and 2D image pixel coordinates $(u, v)$ is given by 

\begin{equation}
s\begin{bmatrix}
u \\
v \\
1
\end{bmatrix} = K [R|t]\begin{bmatrix}
x \\
y \\
z \\
1
\end{bmatrix} = \begin{bmatrix}
p_{11} & p_{12} & p_{13} & p_{14} \\
p_{21} & p_{22} & p_{23} & p_{24} \\
p_{31} & p_{32} & p_{33} & p_{34}
\end{bmatrix}\begin{bmatrix}
x \\
y \\
z \\
1
\end{bmatrix},
\end{equation}

where $s$ is a scaling factor, $P = K [R|t]$ is the $3 \times 4$ perspective transformation matrix, $K$ is the intrinsic camera matrix, and $[R|t]$ is the $3 \times 4$ extrinsic parameter matrix. For our BEV projection, we assume all points lie on the ground plane $(z = 0)$. This simplifies the projection to 

\begin{equation}
s\begin{bmatrix}
u \\
v \\
1
\end{bmatrix} = P'\begin{bmatrix}
x \\
y \\
1
\end{bmatrix} = \begin{bmatrix}
p_{11} & p_{12} & p_{14} \\
p_{21} & p_{22} & p_{24} \\
p_{31} & p_{32} & p_{34}
\end{bmatrix}\begin{bmatrix}
x \\
y \\
1
\end{bmatrix},
\end{equation}

where $P'$ is the $3 \times 3$ perspective transformation matrix without the third column of $P$. We apply this transformation to project features from all $S$ cameras, with their respective projection matrices $P'^{(s)}$, onto a predefined ground plane grid of size $[H_g, W_g]$. Each grid position represents an area of 10 cm $\times$ 10 cm. The resulting BEV feature has a size of $S \times C_f \times H_g \times W_g$, where $C_f$ is the number of feature channels. Note, it is important to distinguish between different types of transformation matrices used in our framework. The perspective projection matrix $P$ is a $3 \times 4$ matrix that projects 3D world coordinates to 2D image coordinates. The homography matrix is always a $3 \times 3$ matrix that maps points between two planes (e.g., from the image plane to the ground plane). Our simplified perspective projection matrix $P'$ is a $3 \times 3$ matrix derived from $P$ by assuming $z=0$. In the context of our BEV projection, $P'$ functions similarly to a homography matrix.

\subsection{BEV Detection and Tracking} \label{detntrack}

The decoder output bifurcates into BEV detection and re-identification branches. The BEV detection branch employs four prediction heads: center prediction (1-channel), offset prediction (2-channel), size prediction (3-channel), and rotation prediction (8-channel). These predictions undergo post-processing, including non-maximum suppression and top-K selection, yielding final BEV detections. Concurrently, the re-identification branch extracts identity features from both BEV and image-level representations.

The tracking module integrates BEV detections with identity features, employing a Joint Detection and Embedding (JDE) tracker. This tracker associates detections across frames using both spatial and appearance information, enabling robust object tracking in the BEV domain despite occlusions and view changes.

\section{Performance Evaluation}~\label{Section_IV}

\begin{table}
 \caption{Simulation Parameters}
 \centering
 \begin{tabular}{l c c}
 	\hline\hline
 	Parameter & Symbol &Value \\
 	\hline
    Height of the original image & $H_\mathrm{o}$ &  720 px  \\
    Width of the original image & $W_\mathrm{o}$ &  1280 px \\
    Height of the resized image & $H_\mathrm{r}$ &  360 px  \\
    Width of the resized image & $W_\mathrm{r}$ &  640 px  \\
    Semantically-guided masking hyperparameter & $\kappa$ & 0.15  \\
    Patch size & 20 $\times$ 20 px \\
    Bits per pixel & 24 \\
 	\hline
 \end{tabular}
 \label{Table_Sim}
\end{table}   

This section presents the performance evaluation of our proposed communication-efficient distributed multiview detection and tracking system using MAEs. We analyze the results obtained from experiments conducted on the MultiviewX and Wildtrack datasets, focusing on detection and tracking metrics, as well as communication efficiency. The simulation parameters used for the experiments are shown in Table~\ref{Table_Sim}. We compare the performance of our proposed system with the state-of-the-art work presented in~\cite{teepe2023earlybird}. The performance of our proposed system is evaluated against the masking ratio $r$, defined as $r = N_\text{masked} / N_\text{total}$, where $N_\text{masked}$ is the number of masked patches and $N_\text{total}$ is the total number of patches in the image. A higher masking ratio indicates that fewer patches are transmitted, resulting in greater data compression but potentially reduced image quality. This ratio quantifies the trade-off between communication efficiency and system performance across different compression levels.

\subsection{MultiviewX Dataset}
The MultiviewX dataset~\cite{hou2020multiview} presents a challenging multiview pedestrian detection and tracking scenario. It comprises synchronized video streams from six calibrated cameras, capturing an outdoor scene with numerous pedestrians in a virtual world generated by a video game engine. The dataset includes 400 frames, with bounding box annotations provided in the ground plane. This dataset is particularly valuable for evaluating algorithms' ability to handle occlusions and varying pedestrian densities in a multiview setup.

\subsection{Wildtrack Dataset}
Wildtrack~\cite{8578626} offers a real-world environment for multiview pedestrian detection and tracking. It features seven synchronized and calibrated cameras recording an outdoor scene. The dataset consists of 400 frames, split evenly between training and testing sets. Ground truth annotations are provided as point annotations on the ground plane for each pedestrian. Wildtrack's diversity in camera angles and pedestrian interactions makes it an excellent benchmark for assessing the robustness of multiview tracking algorithms.

\subsection{Evaluation Metrics}
To comprehensively assess the performance of multiview detection and tracking algorithms, we employ four key metrics, two for detection~\cite{10028728} and two for tracking performance~\cite{bernardin2008evaluating}:

\begin{itemize}
    \item \textbf{MODA (Multiple Object Detection Accuracy):} This metric evaluates the accuracy of detections by considering both false positives and false negatives. It provides an overall measure of detection quality, penalizing both missed detections and false alarms.
    
    \item \textbf{MODP (Multiple Object Detection Precision):} MODP assesses the localization precision of correct detections. It quantifies how well the detected bounding boxes align with the ground truth, offering insight into the spatial accuracy of the detections.
    
    \item \textbf{MOTA (Multiple Object Tracking Accuracy):} This metric extends MODA to the tracking domain. It considers not only false positives and false negatives but also identity switches in tracking. MOTA provides a comprehensive view of a system's ability to track objects over time consistently.
    
    \item \textbf{MOTP (Multiple Object Tracking Precision):} Similar to MODP, MOTP measures the localization precision of tracks. It evaluates how closely the predicted trajectories match the ground truth paths, giving an indication of the tracking system's spatial accuracy over time.
\end{itemize}

These metrics collectively offer a multi-faceted evaluation of both detection and tracking performance, enabling a thorough analysis of multiview pedestrian tracking systems.

\begin{table*}
\centering
\begin{threeparttable}
\caption{State-of-the-Art Performance}
\label{Table_SOTA}
\setlength{\tabcolsep}{4pt}
\begin{tabular}{@{}l*{10}{c}@{}}
\toprule
\multirow{2}{*}[-0.5ex]{Paper} & \multicolumn{5}{c}{Wildtrack} & \multicolumn{5}{c}{MultiviewX} \\
\cmidrule(lr){2-6} \cmidrule(lr){7-11}
& MODA~[\%] & MODP~[\%] & MOTA~[\%] & MOTP~[\%] & \begin{tabular}[c]{@{}c@{}}Comm.\\ Vol.[Mb]\end{tabular} & MODA~[\%] & MODP~[\%] & MOTA~[\%] & MOTP~[\%] & \begin{tabular}[c]{@{}c@{}}Comm.\\ Vol.[Mb]\end{tabular} \\
\midrule
MVTT~\cite{10031058} & \textbf{94.1} & \underline{81.3} & - & - & \multirow{5}{*}{\underline{154.8}} & \underline{95.0} & \textbf{92.8} & - & - & \multirow{5}{*}{\underline{132.7}} \\
MVDet~\cite{hou2020multiview} & 88.2 & 75.7 & - & - & & 83.9 & 79.6 & - & - & \\
MVDeTr~\cite{zhu2021deformable} & \underline{91.5} & \textbf{82.1} & - & - & & 93.7 & \underline{91.3} & - & - & \\
EarlyBird~\cite{teepe2023earlybird} & 91.2 & 81.8 & \underline{89.5} & \underline{86.6} & & 94.2 & 90.1 & \underline{88.4} & \textbf{86.2} & \\
TrackTacular~\cite{teepe2024lifting} & 91.8 & 79.8 & \textbf{89.6} & {81.7} & & \textbf{95.9} & 89.2 & \textbf{91.4} & \underline{86.7} & \\
\midrule
\multirow{3}{*}[-1ex]{\begin{tabular}[c]{@{}l@{}}\textbf{Our method}\\\small{(70\% mask ratio)}\end{tabular}} & 
\multirow{3}{*}{90.9} & \multirow{3}{*}{79.4} & \multirow{3}{*}{88.5} & \multirow{3}{*}{\textbf{86.8}} & \multirow{3}{*}{\textbf{11.7}} & 
\multirow{3}{*}{89.9} & \multirow{3}{*}{90.5} & \multirow{3}{*}{81.0} & \multirow{3}{*}{85.8} & \multirow{3}{*}{\textbf{10.0}} \\
\\
\\[1ex]
\bottomrule
\end{tabular}
\begin{tablenotes}
 \item[1] The measured performance of the EarlyBird system is 1\%-2\% different from the results documented in the paper~\cite{teepe2023earlybird} due to the Python test code behaving differently from the MATLAB code. This table presents the results taken from the respective research work. Elsewhere within our work, the performance is simulated with our local machine.
\end{tablenotes}
\end{threeparttable}
\end{table*}

\begin{figure}[!t]
    \normalsize
	\centering
   	\includegraphics[width=\linewidth]{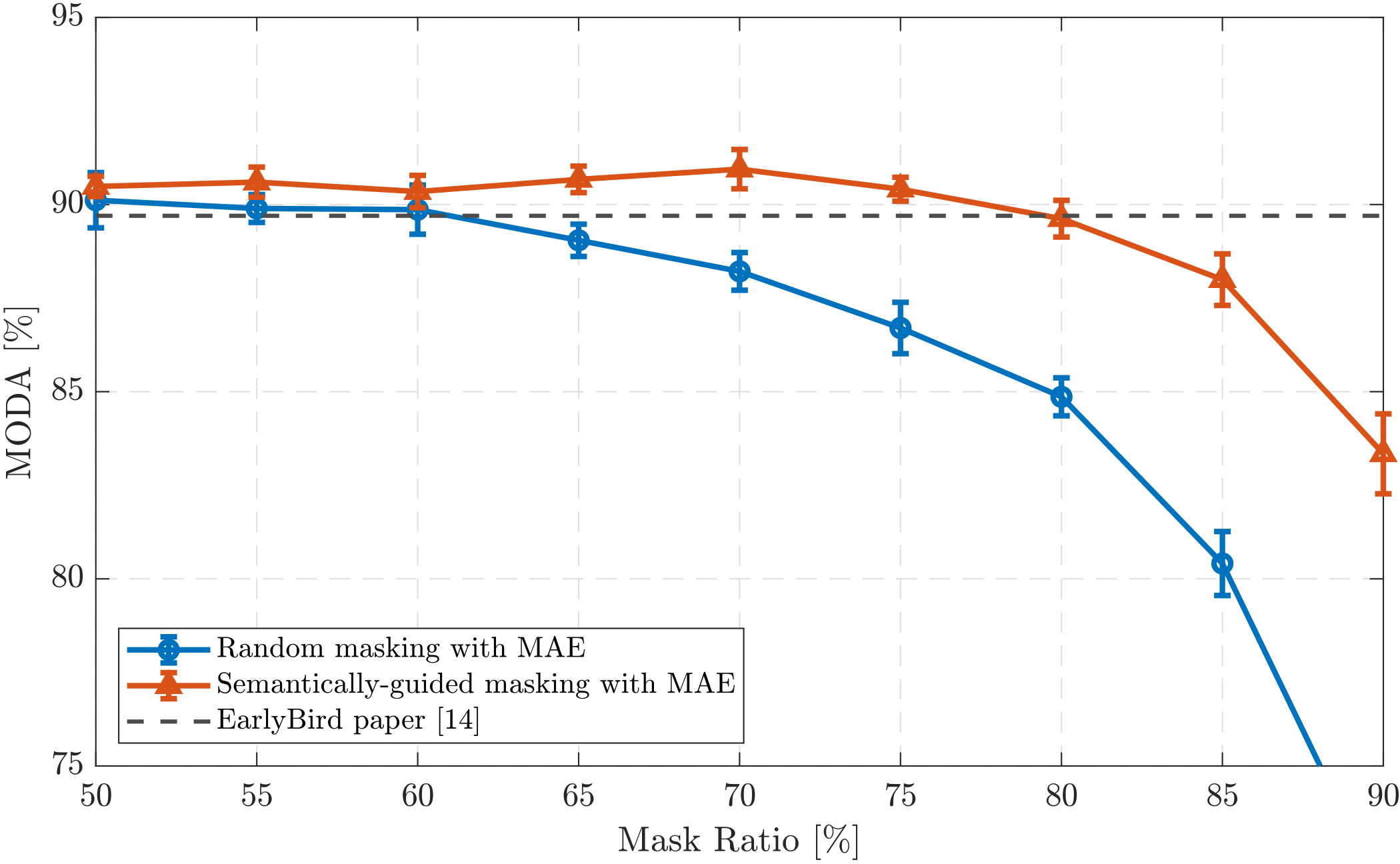}
	\caption{A plot of the MODA as the masking ratio increases for the Wildtrack dataset. For the semantically-guided masking $\kappa = 0.15$.}
	\label{Fig_mod_comp}
\end{figure}

\begin{figure}[!t]
    \normalsize
	\centering
   	\includegraphics[width=\linewidth]{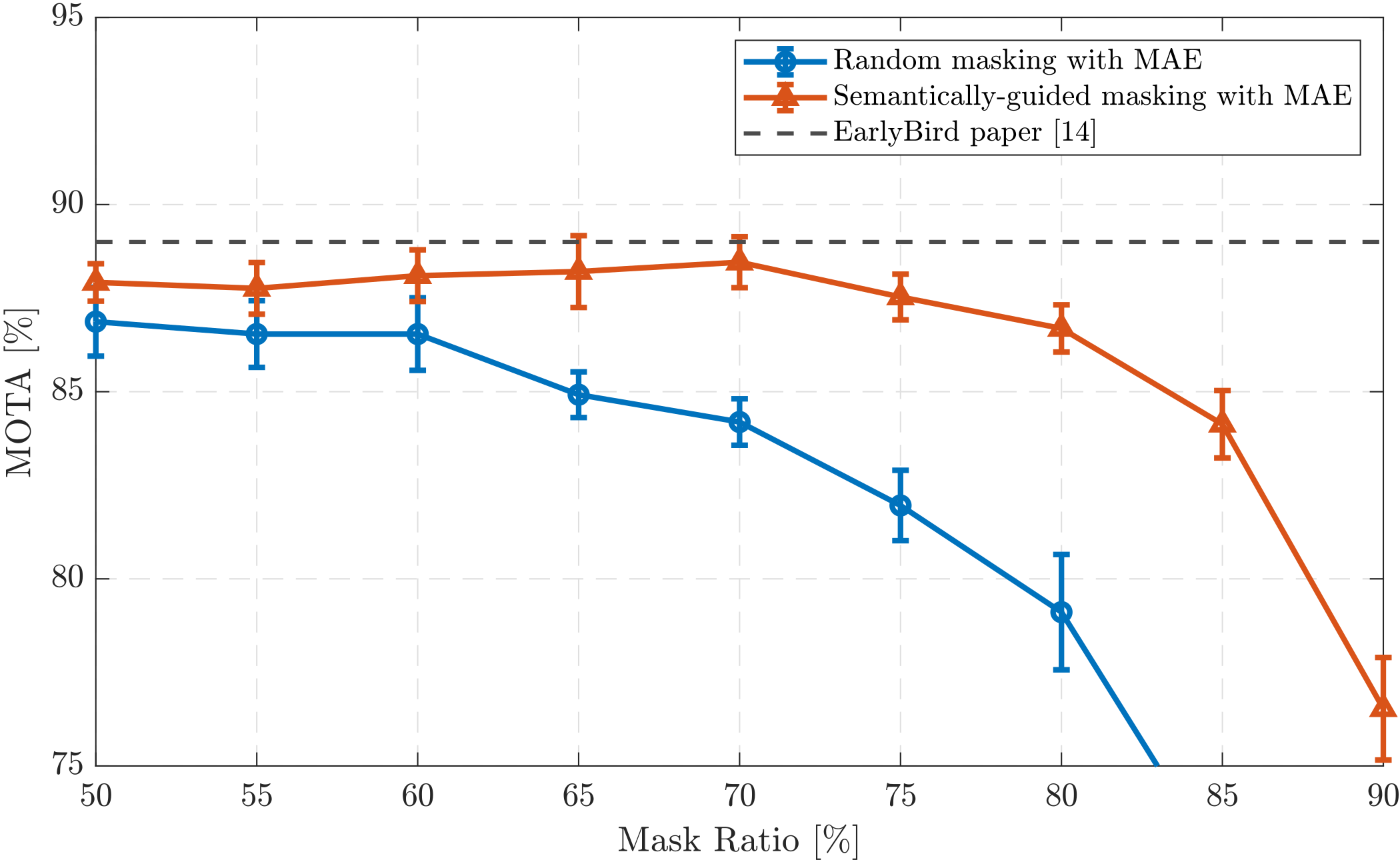}
	\caption{A plot of the MOTA  as the masking ratio increases for the Wildtrack dataset. For the semantically-guided masking $\kappa = 0.15$.}
	\label{Fig_mot_comp}
\end{figure}

\subsection{Detection and Tracking Performance}

We first test the detection performance which is shown in Fig.~\ref{Fig_mod_comp}. From this plot, we can see that the performance of our proposed method closely aligns with the state-of-the-art work presented in~\cite{teepe2023earlybird} while maintaining a strong MODP performance of $\approx$80\%, aligning with the state-of-the-art. Utilizing the semantically-guided masking algorithm yields an even better performance curve than the random masking MAE framework. This is expected since the semantically-guided masking technique focuses on more informative parts of the image. In addition, comparing the accuracy results with and without an MAE we can see that the accuracy performance increases slightly. When using the MAE together with the semantically guided masking algorithm displays even higher performance than the state-of-the-art.

Fig.~\ref{Fig_mot_comp} shows the tracking results which demonstrates similar performance characteristics To the state-of-art also maintaining a strong MOTP performance of $\approx$86\%, Again the accuracy is closely aligned with the state-of-the-art work and utilizing the semantically-guided masking algorithm displays better performance results Compared to the random masking algorithm. However, the accuracy performance does not seem to be affected too much with or without an MAE.

\begin{figure}[!t]
    \normalsize
	\centering
   	\includegraphics[width=\linewidth]{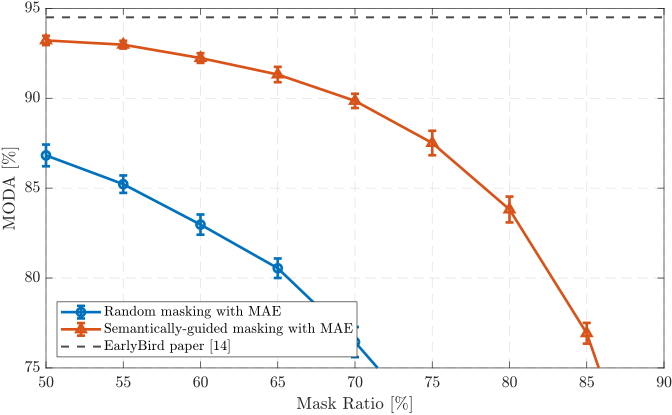}
	\caption{A plot of the MODA as the masking ratio increases for the MultiviewX dataset. For the semantically-guided masking $\kappa = 0.15$.}
	\label{Fig_mod_comp_multiview}
\end{figure}

\begin{figure}[!t]
    \normalsize
	\centering
   	\includegraphics[width=\linewidth]{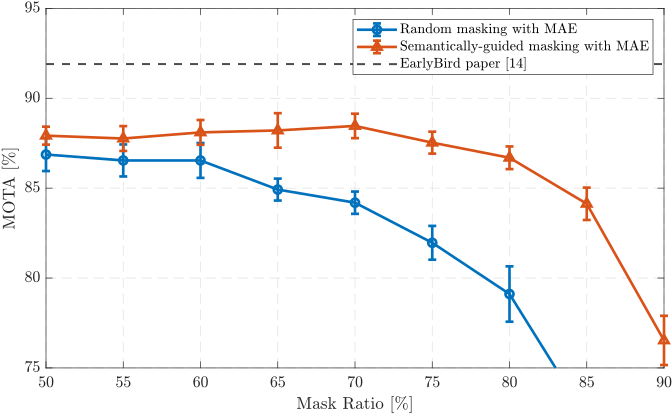}
	\caption{A plot of the MOTA as the masking ratio increases for the MultiviewX dataset. For the semantically-guided masking $\kappa = 0.15$.}
	\label{Fig_mot_comp_multiview}
\end{figure}

Fig.~\ref{Fig_mod_comp_multiview} and Fig.~\ref{Fig_mot_comp_multiview} illustrate the detection and tracking performance, respectively, for the MultiviewX dataset. These results exhibit similar trends to those observed in the Wildtrack dataset analysis but with some notable distinctions.

As shown in Fig.~\ref{Fig_mod_comp_multiview}, the detection performance for the MultiviewX dataset follows a pattern comparable to that of the Wildtrack dataset. However, a more pronounced difference is evident between the semantically-guided masking and random masking techniques. The semantically-guided masking approach demonstrates a higher performance across various masking ratios.

Fig.~\ref{Fig_mot_comp_multiview}, which depicts the tracking results, presents a similar narrative. The performance characteristics align with those observed in the detection results, with the semantically-guided masking algorithm showcasing superior performance even at higher masking ratios. The discrepancy between semantically-guided and random masking is more substantial compared to the Wildtrack dataset results.

We hypothesize that the more substantial performance gap between semantically-guided and random masking in the MultiviewX dataset is attributed to the higher concentration of pedestrian targets. The MultiviewX dataset contains an average of 40 pedestrians per frame, compared to approximately 20 in the Wildtrack dataset~\cite{hou2020multiview, 8578626}. This increased density of targets likely amplifies the benefits of the semantically-guided masking technique.

The higher pedestrian count per frame in MultiviewX means that there are more informative regions in each image for the semantically-guided masking algorithm to focus on. Consequently, the algorithm can more effectively prioritize areas with valuable information, leading to better preservation of critical features even at higher masking ratios. This results in a more robust performance compared to random masking, which does not account for the distribution of informative regions.

\subsubsection{Comparison with State-of-the-Art Methods}

Table~\ref{Table_SOTA} presents a comprehensive comparison of our proposed method with state-of-the-art approaches for multiview detection and tracking. Our method, operating at a 70\% masking ratio, demonstrates competitive performance across all metrics for both the Wildtrack and MultiviewX datasets.

What sets our approach apart is its ability to achieve these competitive results while reducing the communication cost. By employing a 70\% masking ratio, our method transmits only 30\% of the image data compared to other approaches. This reduction in data transmission is further amplified by our image-resizing strategy, which reduces the image dimensions by a factor of two in both height and width. Consequently, our method achieves a 13.33-fold reduction in data volume ($0.3 \times 0.25 = 0.075$, or 1/13.33) compared to methods that transmit full-resolution, unmasked images. Furthermore, to get the communication volume over time we can multiply the data usage by two as both datasets operate at two frames per second.



\subsection{Dropout}

\begin{figure}[!t]
    \normalsize
	\centering
   	\includegraphics[width=\linewidth]{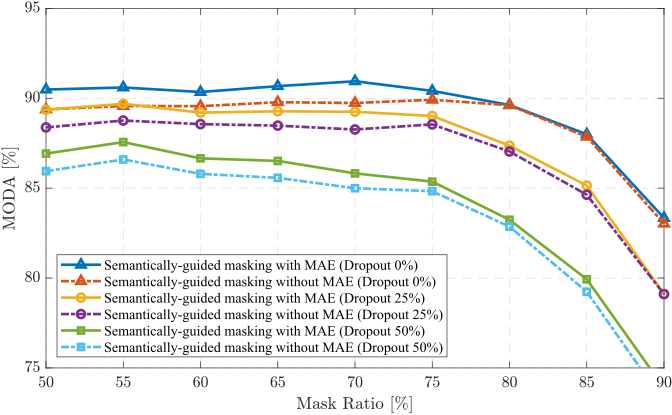}
	\caption{A plot of the MODA performance for different camera dropout percentages on the Wildtrack dataset.}
	\label{Fig_dropout}
\end{figure}

Fig.~\ref{Fig_dropout} demonstrates the effect of camera dropout on MODA performance. As dropout percentage increases from 0\% to 50\%, MODA decreases for both pipeline configurations. Notably, the performance gap between the semantically guided with the MAE pipeline and the pipeline without MAE remains consistent across all dropout rates.

This consistency suggests that the MAE component provides a robust performance improvement, maintaining its advantage even as camera availability decreases. The MAE-enhanced pipeline's resilience to camera dropout indicates its potential value in real-world scenarios where camera failures or obstructions are common.

\section{Conclusion}~\label{Section_V}

In this work, we presented a novel approach for communication-efficient distributed multiview detection and tracking using masked autoencoders. Our method addresses the critical challenges of bandwidth limitations and computational constraints in multiview systems, particularly for resource-limited camera nodes. By leveraging a semantic-guided masking strategy and MAE-based reconstruction, our approach reduces communication overhead while preserving essential visual information for accurate detection and tracking. The integration of a pre-trained semantic segmentation model and a tunable power function allows for adaptive masking that prioritizes the most informative image regions.

Our extensive experiments on the MultiviewX and Wildtrack datasets demonstrate the effectiveness of our approach. The results show that our method achieves comparable or better performance in terms of MODA, MODP, MOTA, and MOTP metrics compared to state-of-the-art techniques, even at high masking ratios. Notably, our semantically-guided masking algorithm outperforms random masking, maintaining higher accuracy and precision as the masking ratio increases. Furthermore, our approach achieves significant communication efficiency, reducing data volume by leveraging a masking technique to efficiently send over only the useful information and an MAE for reconstruction. This efficiency gain is crucial for real-world deployments with limited bandwidth and energy constraints. Future work will investigate communication-efficient multiview detection and tracking in non-terrestrial networks.

\section*{Acknowledgements}

We thank the Defence Innovation Network and NSW State Government for financial support of this project through grant DIN Pilot Project grant 2023-24. The authors thank Saksham Yadav, founder of DeepNeural AI, for his participation in project meetings as an industry contact.

\bibliographystyle{ieeetr}
\bibliography{bibliography}

\clearpage

\end{document}